\documentclass[letterpaper]{article} 
\usepackage{aaai25}  
\usepackage{times}  
\usepackage{helvet}  
\usepackage{courier}  
\usepackage[hyphens]{url}  
\usepackage{graphicx} 
\usepackage{natbib}  
\usepackage{caption} 
\usepackage{makecell}
\usepackage{algorithm}
\usepackage{algorithmic}

\usepackage{newfloat}
\usepackage{listings}
\DeclareCaptionStyle{ruled}{labelfont=normalfont,labelsep=colon,strut=off} 
\floatstyle{ruled}
\newfloat{listing}{tb}{lst}{}
\floatname{listing}{Listing}
\usepackage{color} 

\title{AutoSGNN: Automatic Propagation Mechanism Discovery for Spectral Graph Neural Networks}
\author{
    Shibing Mo\textsuperscript{\rm 1,2},
    Kai Wu\textsuperscript{\rm 1},
    Qixuan Gao\textsuperscript{\rm 2},
    Xiangyi Teng\textsuperscript{\rm 2}\thanks{Corresponding author.},
    Jing Liu\textsuperscript{\rm 1,2}
}
\affiliations{
    \textsuperscript{\rm 1}School of Artifcial Intelligence, Xidian University\\
    \textsuperscript{\rm 2}Guangzhou Institute of Technology, Xidian University

    msb@stu.xidian.edu.cn, kwu@xidian.edu.cn, gaoqx@stu.xidian.edu.cn, tengxiangyi@xidian.edu.cn, neouma@mail.xidian.edu.cn
}

\usepackage{bibentry}

\begin{document}

\maketitle

\begin{abstract}
In real-world applications, spectral Graph Neural Networks (GNNs) are powerful tools for processing diverse types of graphs. However, a single GNN often struggles to handle different graph types—such as homogeneous and heterogeneous graphs—simultaneously. This challenge has led to the manual design of GNNs tailored to specific graph types, but these approaches are limited by the high cost of labor and the constraints of expert knowledge, which cannot keep up with the rapid growth of graph data. To overcome these challenges, we propose AutoSGNN, an automated framework for discovering propagation mechanisms in spectral GNNs. AutoSGNN unifies the search space for spectral GNNs by integrating large language models with evolutionary strategies to automatically generate architectures that adapt to various graph types. Extensive experiments on nine widely-used datasets, encompassing both homophilic and heterophilic graphs, demonstrate that AutoSGNN outperforms state-of-the-art spectral GNNs and graph neural architecture search methods in both performance and efficiency.
\end{abstract}

%
\begin{links}
    \link{Code}{https://github.com/Explorermomo/AAAI2025-AutoSGNN}
\end{links}

\section{Introduction}

Graph Neural Networks (GNNs) have become central to machine learning, with recent advancements dividing them into spatial-based \cite{10} and spectral-based \cite{8} approaches. Spectral-based methods focus on creating advanced spectral graph filters for effective graph embedding, and have a explicit and explainable propagation mechanism for graphs, effectively characterizing the key features during the learning process. 

\begin{figure}[t]
\centering
\includegraphics[width=1.0\columnwidth]{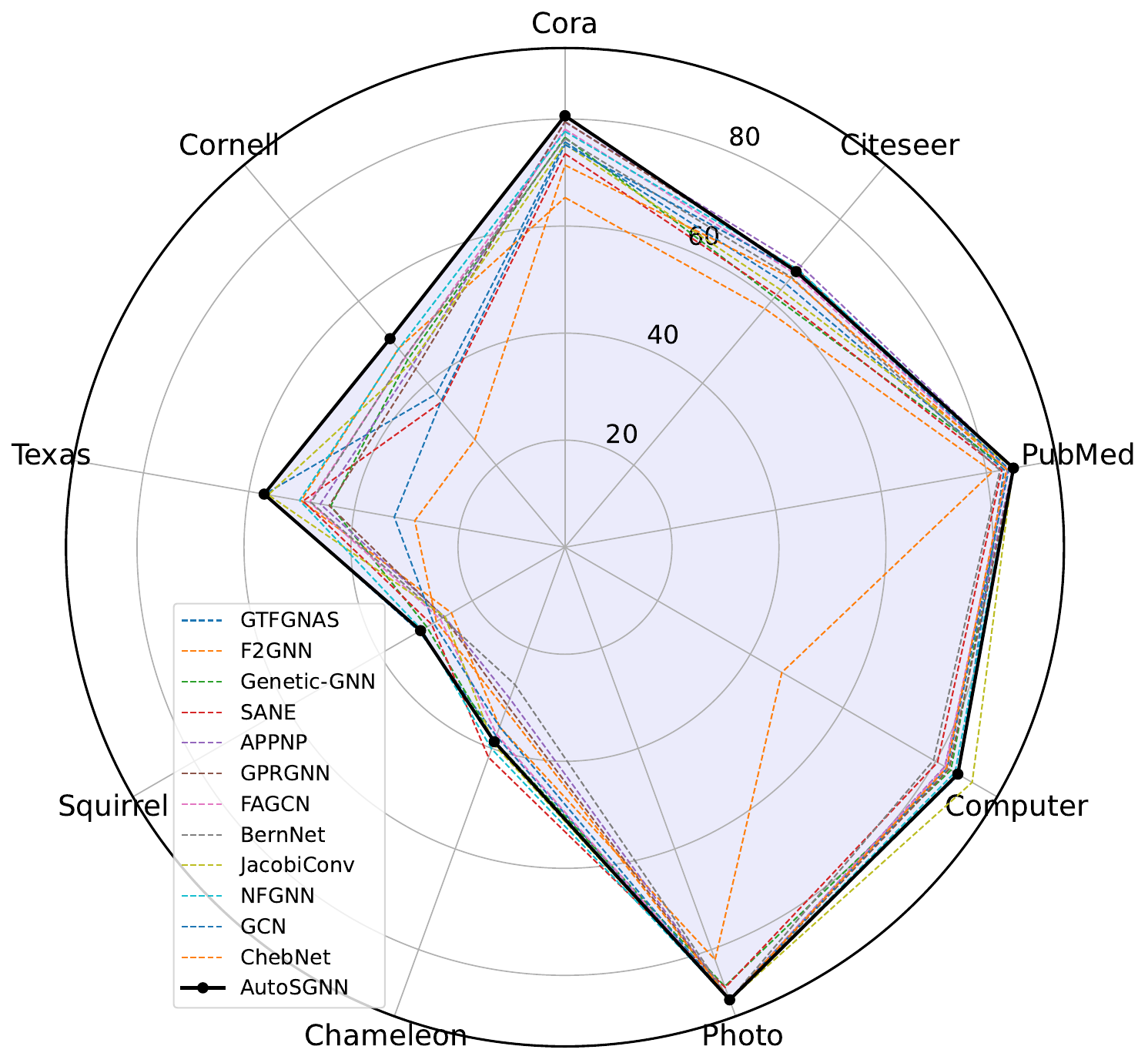} 
\caption{Radar chart of AutoSGNN and other spectral GNNs on 9 datasets. }
\label{leida}
\end{figure}

Many traditional spectral GNNs assume homogeneity \cite{12,13}, where interconnected nodes typically belong to the same category or share similar attributes. However, real-world graphs often exhibit significant heterogeneity, such as protein structure networks \cite{14} that show stronger heterophilic connections between different types of amino acids. GNNs designed with the homogeneity assumption may struggle to perform well on such heterogeneous graphs \cite{15,16}. Thus, the generalization ability of spectral GNNs is tested, and we find that a single spectral GNN is difficult to adapt to different graph data (see Figure \ref{leida}, the advantageous application scenarios of different neural networks have differences). However, the current different spectral GNNs are manually designed to adapt to particular scenarios. Automatic and efficient GNN configuration schemes are urgently needed to reduce labor costs. 

To solve this issue, in the past five years, many Neural Architecture Search (NAS) methods specifically designed for GNN neural architecture search (GNN-NAS) have been proposed \cite{18,19}.  GNN-NAS can be summarized as a combinatorial optimization problem. Limited by the quantity and quality of SGNNs designed by current experts, it is difficult to adapt to a wider range of application scenarios. Additionally, while many efforts have been made to design adaptive spectral GNNs  \cite{26,27}, their generalization capability remains constrained by issues such as homogeneity assumptions and oversmoothing problem.

Recently, Large Language Models (LLMs) have garnered significant attention for their powerful capabilities and versatility in solving natural language processing tasks such as text generation and question answering. Studies such as \cite{20,21,22} have demonstrated that LLMs can be used to discover scientific equations and mathematical algorithms, providing opportunities for numerous fields, including graph learning.

In this paper, we propose AutoSGNN, a framework for the automatic discovery of spectral GNNs. By leveraging LLMs, AutoSGNN can identify or generate suitable spectral GNNs tailored to various conditions. Specifically, we design a search space for spectral GNNs that encompasses the feature fitting terms, the graph Laplacian regularization terms, and the aggregation terms. This search space is described through a combination of text and code, enabling LLMs to understand and generate appropriate spectral GNNs. Then, based on a unified spectral GNN architecture, we establish an evolutionary framework to collaboratively evolve the design thoughts and code of spectral GNNs. With the assistance of LLMs, the linguistic description and code of heuristics evolve iteratively, guided by curated prompts, to generate the most suitable spectral GNNs for different graphs. In summary, AutoSGNN distinguishes itself from previous works through some key features:
\begin{itemize}
\item AutoSGNN introduces an automated approach for designing feature fitting terms, graph Laplacian regularization terms, and aggregation terms in spectral GNNs. This method is capable of generating suitable spectral GNNs for a wide range of graph datasets that encompass both homogeneous and heterogeneous graphs.
\item We propose the fourth type of method for GNN-NAS, distinct from existing three types of approaches that rely on gradient descent, reinforcement learning, or evolutionary algorithms. AutoSGNN achieves a high degree of parallelism throughout the entire process, from generation to evaluation.
\item In contrast to GNN-NAS methods, which are essentially combinatorial optimization techniques, AutoSGNN goes beyond by truly generating GNN frameworks through learning and understanding the fundamental design principles of filters.
\item We conducted a comprehensive evaluation of AutoSGNN across nine widely studied graph datasets. Extensive experimental results demonstrate that AutoSGNN exhibits exceptional competitiveness in both algorithmic performance and time complexity, outperforming or matching state-of-the-art spectral GNNs and GNN-NAS methods.
\end{itemize}

\section{Related Work}
\subsubsection{Spectral Graph Neural Network}
Spectral GNNs design spectral graph filters in the spectral domain. ChebNet \cite{31} employs higher-order Chebyshev polynomials to approximate the filters. GCN \cite{24} uses a first-order approximation of the Chebyshev filters to simplify the process. GPRGNN \cite{26} approximates polynomial filters through gradient descent on the polynomial coefficients. FAGCN \cite{25} learns the weights of low-pass and high-pass filters by approximating them with attention coefficients. BernNet \cite{27} approximates any non-negative coefficient filter in the normalized Laplacian spectrum of the graph using k-order Bernstein polynomials. NFGNN \cite{9} creates a local filter by combining the Kronecker delta function with polynomial approximations, and accelerates computational efficiency while balancing local and global spectral information by reparameterizing the feature parameters.

Despite the emergence of various spectral GNNs, their graph learning mechanisms can mostly be viewed as optimizing the feature fitting functions of various graph kernels with graph regularization terms. \cite{23} optimizes two proposed objective functions, allowing adjustable high-pass or low-pass filters to fit a wide range of GNNs.
\subsubsection{Neural Achitecture Search}
Recently, NAS has been applied to GNNs to automate their design process. GraphNAS \cite{32} and AutoGraph \cite{33} utilize reinforcement learning and genetic algorithms, respectively, to find the optimal GNN architecture. Recent methods like SANE \cite{19} employ differentiable NAS, reducing computational costs through gradient-based optimization. CTFGNAS \cite{18} introduces an alternative evolutionary Graph NAS (GNAS) algorithm, whose search space includes not only micro-level network layer components but also topological connections and feature fusion strategies. These approaches demonstrate that NAS has the potential to enhance GNN performance across various tasks.
\subsubsection{LLM for Graph}
In recent years, LLMs such as ChatGPT \cite{35}  have been applied to fields like mathematical formula discovery due to their outstanding text processing and generation capabilities. \cite{21} applys LLMs to heuristic algorithm discovery. By adding heuristic prompts that approximate the chain of thought, \cite{20} can discover heuristic algorithms more rapidly. Meanwhile, \cite{22} utilized LLMs to discover scientific equations. The inherent rich domain knowledge and contextual understanding of LLMs enable LLM-based search architectures to be more intelligent, flexibly generating suitable architectures based on different datasets and task requirements. Additionally, Evolutionary Strategy (ES) can provide an optimization framework \cite{kaige1,kaige2,kaige3,kaige4} for further enhancing LLMs in a black-box setting, endowing LLMs with flexible global search capabilities.

\begin{figure*}[t]
\centering
\includegraphics[width=0.71\textwidth]{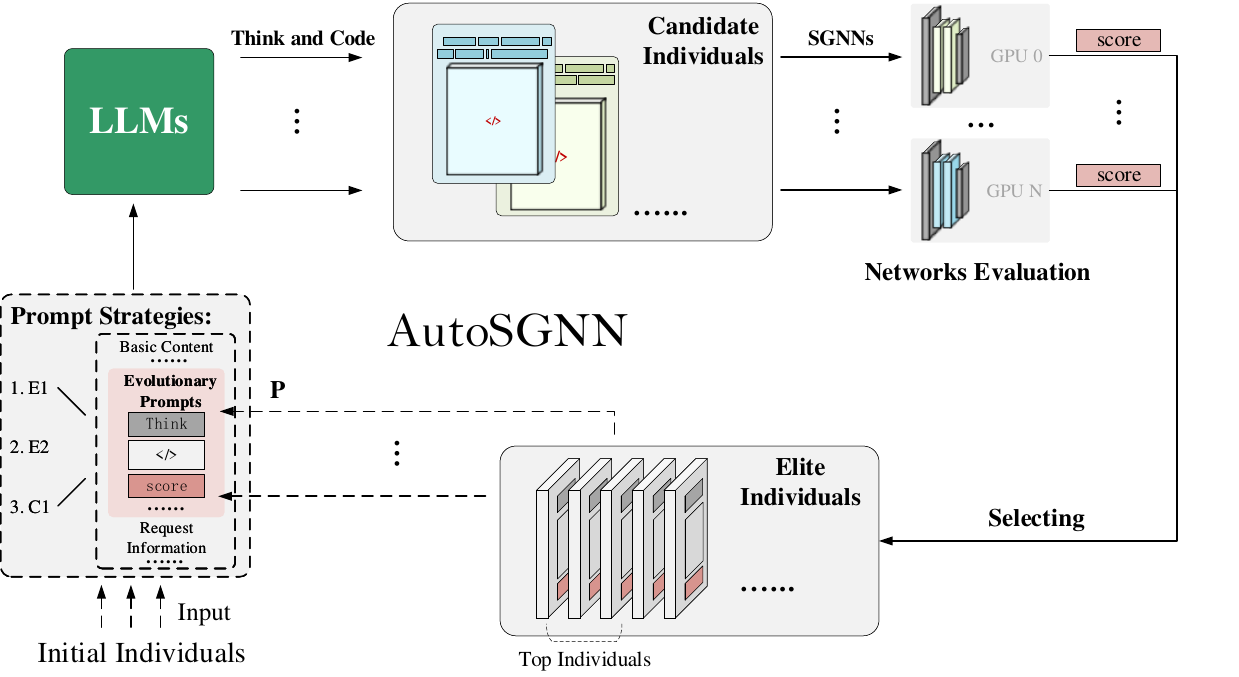} 
\caption{The overview of AutoSGNN. In this process, the LLM is used to generate population individuals, while other components are responsible for evaluation, selection, and describing requirements to optimize the LLM preferences, thus achieving the overall optimization process.}
\label{fig1}
\end{figure*}

\section{Preliminaries}
\subsubsection{Notations}
A graph can be represented as $G=(V,E,A)$, where $V=\{\nu_{i}\}_{i=1}^{n}$ denotes the set of nodes with $\left|V\right|=n$, $E$ is the set of edges between nodes, and the topology of the graph $G$ can be represented by the adjacency matrix $A\in R^{n\times n}$. When $A_{ij}=1$, it means that node $v_i$ is connected to node $v_j$; otherwise $A_{ij}=0$. The features of each node can be represented as $X\in R^{n\times f}$, where $f$ is the node feature dimension. The pairwise angle matrix is denoted as $D=diag(d_1,...,d_n)$, where $d_{i}=\sum_{j}A_{i,j}$.

\subsubsection{Graph Fourier Transform}
The (combinatorial) graph Laplacian is defined as $L=D-A$, which is Symmetric Positive Semi-Definite (SPSD). Its eigendecomposition gives $L=U\Lambda U^{T}$, where the columns $u_i$ of $U\in R^{n\times n}$ are orthonormal eigenvectors, namely the graph Fourier basis, $\Lambda=\mathrm{diag}(\lambda_1,\ldots,\lambda_n)$, where $\lambda_{1}\leq\cdots\leq\lambda_{n}$, and these eigenvalues are also called frequencies. Therefore, the graph Fourier transform of signal $x$ can be defined as $\tilde{x}=U^{-1}x=U^{T}x$, while the inverse Fourier transform is $x=U\tilde{x}$, and the transform of the signal $x$ through filter $\tilde{g}=f(\lambda)=diag(f(\lambda_1),\ldots,f(\lambda_n))$ can be expressed as $z=U\tilde{g}U^{T}x$. In addition to $L$, there exist some common variants of this graph Laplacian matrix such as the symmetric normalized Laplacian matrix $L_{sym}=I-D^{-1/2}AD^{-1/2}$, and random walk normalized Laplacian Matrix $L_{rw}=I-D^{-1}A$. Applying the renormalization trick to affinity and Laplacian matrices respectively leads to $\tilde{L}_{sym}=I-\tilde{D}^{-1/2}\tilde{A}\tilde{D}^{-1/2}$ and $\tilde{L}_{rw}=I-\tilde{D}^{-1}\tilde{A}$, where $\tilde{A}=A+I$, $\tilde{D}=D+I$. In addition to this, \cite{9} uses $\tilde{L}=\frac{2L_{sym}}{\lambda_{\mathrm{max}}}-I$ to limit the range of eigenvalues in order to satisfy numerical stability. For the processing of graph Laplacian matrices, it is often necessary to consider the structure of the graph, signal processing requirements and other conditions in order to select a suitable form.

\subsubsection{Spectral Graph Neural Networks}
Spectral-based GNNs aim to learn specific spectral filters given a graph structure and node labels. By optimizing the optimal solution of feature fitting functions of various graph kernels with a graph regularization term, they retain appropriate frequency components for downstream tasks. By summarizing spectral GNNs, \cite{23} proposed two unified frameworks for spectral GNNs. For GNNs with layer-wise feature transformation (e.g., GCN), the output of the $Kth$ layer directly uses the output of the $(K-1)th$ layer. The learning process can be represented as:
\begin{equation}
\label{1}
Z^K =\left\langle Trans(Agg\{G;Z^{(K-1)}\})\right\rangle_{K}
\end{equation}
where $Z^K$ is the graph embedding of the $Kth$ layer, and $Z^{(0)}=X$. Whereas for some networks that separate $Trans()$ and $Agg\{\}$ (e.g., FAGCN \cite{25}), the $Kth$ layer may perform operations such as pooling, concatenation, or attentional aggregation on the outputs of the $(K-1)th$ layer, and then the learning process can be represented as:
\begin{equation}
\label{2}
Z^K =\left\langle Agg\{G;Z^{(K-1)}\}\right\rangle_{K}
\end{equation}
where $Z^{(0)}=Trans(X)$, and $Trans()$ is the feature transformation operation at the corresponding layer, including the non-linear activation function and the learnable weight matrix $W$ of the specific layer. $Agg\{\}$ means aggregating the $(K-1)$ layer output $Z^{K-1}$ along graph $G$ for the $K$th learning operation according to the set spectral filter. The operator $\left\langle\right\rangle_{K}$ depends on specific GNNs and represents the generalized combinatorial operation after $K$ convolutions. 


By simplifying and integrating Eqs.(\ref{1}) and (\ref{2}), a spectral GNNs unified architecture can be obtained:
\begin{equation}
\label{3}
Z^K =\left\langle TRANS(AGG\{G;X^{raw};Z^{(K-1)}\})\right\rangle_{K}
\end{equation}
where $X^{raw}=Linear(X^{(0)})$. $AGG\{\}$ can also combine operations such as pooling, concatenation or attention aggregation, and $TRANS()$ can combine dimensional scaling factors such as $Linear()$. In summary, it is not difficult to see that when suitable graph regularization terms, spectral filter approximation terms, and aggregation terms are determined, a spectral GNN can be obtained.

\section{AutoSGNN}

\begin{figure}[t]
\centering
\includegraphics[width=0.6\columnwidth]{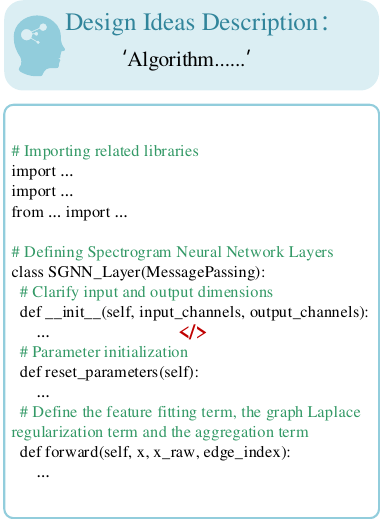} 
\caption{Individual Content Schematic. }
\label{fig2}
\end{figure}
Figure \ref{fig1} shows an overview of AutoSGNN, consisting of five parts: Elite Individuals, Candidate Individuals, Prompt Strategies, LLMs and Networks Evaluation.

\subsection{Elite and Candidate Individuals}
Spectral graph learning mechanisms can mostly be seen as optimizing the feature fitting functions of various graph kernels with graph regularization terms. Therefore, in AutoSGNN, the LLM is used to search for suitable graph regularization, feature approximation terms, and aggregation terms for different graphs, thus generating the most appropriate spectral GNNs.

We use code text descriptions to represent spectral GNNs. Through the formatted output of the LLMs, candidate individuals in the entire evolutionary framework are formed, as shown in Figure \ref{fig2}. The $N$ individuals in candidate individuals, each consist of two parts: the design ideas and the code of the spectral GNNs (containing the feature fitting terms, graph Laplacian regularization terms, and aggregation terms). The spectral GNNs is specifically represented as a Python class, corresponding to the unified architecture of spectral GNNs in Eq.(\ref{3}). By initializing the network's input and output dimensions, it facilitates message passing between layers, and the Design Ideas Description provides a brief introduction to the neural network.

\begin{figure}[t]
\centering
\includegraphics[width=1\columnwidth]{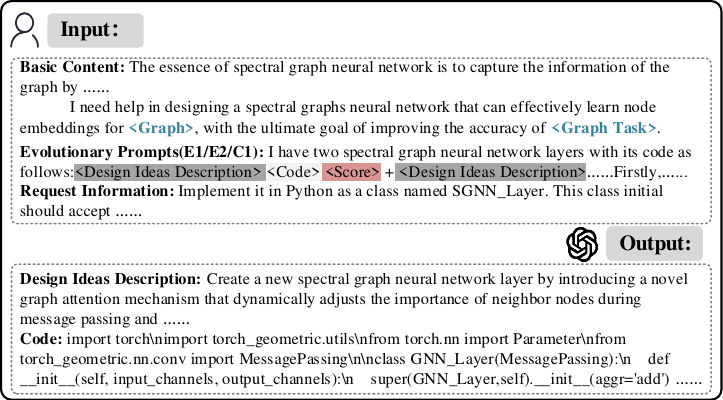} 
\caption{Prompt Strategies for LLMs.}
\label{fig3}
\end{figure}

\begin{table*}[t]
\centering
\caption{The accuracy of node classification under the sparse splitting ratios of 2.5\%:2.5\%:95\%, with \textbf{the best results} highlighted in bold black and \underline{the second-best results} underlined. The results are ranked by employing the Wilcoxon-Holm analysis \cite{IsmailFawaz2018deep} at a significance level of $p=0.05$. }
\label{tab:table 2}
\resizebox{2.1\columnwidth}{!}{
\begin{tabular}{cccccc|cccc|c}
\hline
\textbf{}            & \textbf{Cora}       & \textbf{Citeseer}   & \textbf{PubMed}     & \textbf{Computer}   & \textbf{Photo}      & \textbf{Chameleon}  & \textbf{Squirrel}   & \textbf{Texas}      & \textbf{Cornell}  & \textbf{Rank}   \\ \hline
\textbf{GTFGNAS}     & 75.72±2.06          & 64.24±1.85          & 83.35±1.41          & \underline{84.98±1.41} & 90.52±0.96          & 37.54±5.84          & \textbf{31.47±1.73} & \underline{57.03±3.47}    & 37.38±2.23      & \textbf{4}    \\
\textbf{F2GNN}       & 65.33±1.26          & 58.16±2.28          & 81.16±0.69          & 46.81±4.01          & 82.03±3.13          & 35.72±1.86          & 24.52±1.41          & 50.06±4.47          & \underline{48.44±7.58}   & \textbf{13}   \\  
\textbf{Genetic-GNN} & 76.58±1.06          & 60.98±1.67          & 84.01±0.46          & 83.30±2.10               & 87.33±1.37          & 38.54±2.00          & 29.81±1.09               & 44.45±6.34          & 45.74±7.87       & \textbf{8}   \\
\textbf{SANE}        & 73.52±3.58          & 61.57±2.62          & 82.91±0.26          & 80.41±2.15          & 87.56±1.70          & \textbf{41.79±2.33} & 28.79±1.90          & 49.71±2.92          & 35.61±13.32       & \textbf{11}   \\ \hline
\textbf{APPNP}       & \underline{79.41±0.38}    & \textbf{68.59±0.30} & 85.02±0.09          & 81.99±0.26          & 91.11±0.26          & 30.12±0.92          & 26.01±0.48          & 46.44±3.10          & 44.37±4.90        & \textbf{7}    \\
\textbf{GPRGNN}      & 79.51±0.36          & 67.63±0.38          & \underline{85.07±0.09}    & 82.90±0.37          & 91.93±0.26          & 31.46±0.89          & 26.14±0.51          & 44.76±4.33          & 43.45±4.65        & \textbf{5}   \\
\textbf{FAGCN}       & 78.10±0.21          & 66.77±0.18          & 84.09±0.02          & 82.11±1.55          & 90.39±1.34          & 37.24±3.54          & 26.91±3.08          & 48.44±1.78          & 46.38±1.82        & \textbf{6}  \\
\textbf{BernNet}     & 76.37±0.36          & 65.83±0.61          & 82.57±0.17          & 79.57±0.28          & 91.60±0.35          & 27.33±1.14          & 26.42±0.39          & 48.21±3.17          & 46.50±5.57       & \textbf{10}   \\
\textbf{JacobiConv}     & 75.30±0.02          & 62.75±0.02          & 85.05±0.01          & \textbf{87.89±0.01}      & \underline{92.34±0.01}          & 39.13±0.02          & 25.91±0.01          & 56.34±0.05          & 44.80±0.04       & \textbf{3}   \\
\textbf{NFGNN}       & 77.69±0.91          & \underline{67.74±0.52}    & \underline{ 85.07±0.13}    & 84.18±0.40          & {92.16±0.82}    & \underline{ 40.10±1.34}    & 30.91±0.62          & 50.37±6.54          & 48.37±7.34         & \textbf{2}   \\
\textbf{GCN}         & 75.21±0.38          & 67.30±0.35          & 84.27±0.01          & 82.52±0.32          & 90.54±0.21          & 35.73±0.87          & 28.37±0.77          & 32.40±2.31          & 35.90±3.64        & \textbf{9}   \\
\textbf{ChebNet}     & 71.39±0.51          & 65.67±0.38          & 83.83±0.12          & 82.41±0.28          & 90.09±0.28          & 32.36±1.61          & 27.76±0.93          & 28.51±4.03          & 26.21±3.12       & \textbf{12}   \\ \hline
\textbf{AutoSGNN}    & \textbf{80.61±1.52} & 67.23±1.31          & \textbf{85.13±0.26} & {84.80±0.82}    & \textbf{92.51±0.36} & 38.67±3.52          & \underline{31.17±1.82}    & \textbf{57.10±4.86} & \textbf{50.87±6.50}   & \textbf{1} \\ \hline
\end{tabular}}
\end{table*}

\subsection{Prompt Strategies}
Although LLMs have been widely applied, their black-box nature and limited search capabilities hinder their performance in complex optimization problems. Evolution Strategies(ES), a classic black-box optimization technique \cite{28}, can provide an optimization framework for LLMs in black-box environments, endowing LLMs with flexible global search capabilities. AutoSGNN combines the advantages of ES and LLMs, designing three types of evolutionary prompts(E1, E2, C1) to quickly understand the preferences of different datasets, enabling the selection or generation of optimal spectral GNNs. As shown in Figure \ref{fig3}, the three types of prompts consist of basic information, evolutionary strategies, and requirement information. Specific prompts for different datasets are detailed in \textbf{Appendix B}.

\subsubsection{Basic Content}
A short introduction to the application of spectral GNNs, including the information of graph and downstream tasks. Some expert knowledge can be provided by setting up tips.

\subsubsection{Evolutionary Prompts}
\begin{itemize}
\item $E1$. Exploring new spectral GNN that are completely different from the given spectral GNN. First, select $P_1$ elite individuals from the Elite Individuals, then prompt the LLM to design a spectral GNN different from the chosen elite individuals, accompanied by a brief textual description. It's like the mutation operation in an evolutionary algorithm.
\item $E2$. Based on the observation of the given elite individuals, new spectral GNN is explored. First, $P_2$ individuals are selected from elite individuals. Then, the LLM is instructed to identify the common ideas of the existing spectral GNNs. Finally, a new spectral GNN is designed, based on these common ideas, but differing from the existing spectral GNNs by introducing new elements, along with a brief textual description. It's like the crossover operation in an evolutionary algorithm.

For E1 and E2, we need to select $P_1$ and $P_2$ individuals to form the prompts. We randomly select 50\% individuals from the top 20\% based on scores, and 50\% from the remaining individuals. This approach helps AutoSGNN maintain the optimization trend.

\item $C1$. By comparing individuals, LLM preferences are set. Inspired by algorithms such as DPO \cite{30}, two elite individuals are selected for comparative learning to directly set LLM preferences. First, two individuals are randomly chosen from the top $1/3$ and bottom $1/3$ by score within Elite Individuals, respectively. Then, the LLM is asked to compare their similarities and differences and hypothesize why the higher-score individual is superior. Finally, a more optimal spectral GNN is designed, accompanied by a brief textual description.
\end{itemize}

\subsubsection{Request Information}
The request information supplements the initialization of the generated spectral GNNs, including input dimensions, output dimensions, input variables, output variables, and the end-of-code marker, "Do not give additional explanations." By specifying these requirements to the LLM, the basic framework of the returned code is constrained, improving the executability of the generated spectral GNNs.

\subsection{LLMs}

In AutoSGNN, a wide variety of LLMs, such as ChatGPT \cite{35}, can be flexibly integrated. These LLMs can be open-source models (e.g., GPT-4 or the LLaMA series), proprietary commercial products, or variants fine-tuned for specific tasks or domains. This flexibility allows AutoSGNN to fully leverage the powerful capabilities of LLMs in generating, reasoning, and optimizing GNN architectures.

\subsection{Networks Evaluation}
By evaluating the score of candidate individuals on different datasets, it helps to adjust the LLMs to generate preferences for different situations. The scoring metric can be selected as downstream task metrics such as classification accuracy and loss value. After scoring the candidate individuals, an elite strategy is used to select individuals, retaining a certain number of individuals into Elite Individuals, which are initialized by selecting some classic methods of spectral GNNs \cite{24,25,26,27,9}.

The entire execution process of AutoSGNN is highly parallel, from the parallel responses of the LLM to the parallel evaluation of each GNN model composed of the generated spectral GNNs on different GPUs. Furthermore, to enhance efficiency, we set a timeout duration ($Timeout \  Duration = 600s$). If an SGNN fails to complete its training and testing within this timeframe, the process is terminated, and the SGNN individual is discarded.

\begin{table*}[t]
\centering
\caption{The summary results of the spectral graph neural networks searched by AutoSGNN for different graphs for node classification task under the sparse splitting ratios of 2.5\%:2.5\%:95\%.}
\label{tab:table sum}
\resizebox{1.8\columnwidth}{!}{
\begin{tabular}{c|c|c}
\hline
\textbf{Graph}     & \textbf{Propagation Mechanism} & \textbf{Relevant Parameter} \\ \hline
\textbf{Cora}                 & \makecell[l]{$Z_{0}=\alpha X$   \\ $Z_{k}=Z_{k}+\sum_{k=1}^{K-1}\alpha(1-\alpha)^{k}\hat{A}Z_{k-1}W_{k}$ \\ $Z_{K}=(1-\alpha)^{K}\hat{A}Z_{K-1}W_{K}+Z_{K-1}$ }                            & \makecell[l]{$Initial\   \alpha=0.15, K=4$ \\ $\hat{A}=\tilde{D}^{-1/2}\tilde{A}(\tilde{D}^{T})^{-1/2}, \tilde{A}=A+4I, \tilde{D}=D+4I$}           \\ \hline
\textbf{Citeseer}               &  \makecell[l]{$Z_{0}=Att\cdot X$ \\ $Z_{K}=Z_{0}+\sum_{k=1}^{K-1}(Att\cdot\hat{A}Z_{k-1}W_{k}+0.2X_{raw})$}                             & \makecell[l]{$Initial Att\in R^{1\times n}, K=4$ \\ $\hat{A}=\tilde{D}^{-1/2}\tilde{A}(\tilde{D}^{T})^{-1/2}, \tilde{A}=A+2I, \tilde{D}=D+2I$ }                        \\ \hline
\textbf{PubMed}                   & \makecell[l]{$Z_{0}=\alpha X+\gamma X_{raw}$  \\  $Z_{K}=Z_{0}+\sum_{k=1}^{K}\beta\cdot ReLU(\overline{A}Z_{k-1}W_{k})$}                          & \makecell[l]{$Initial\  \alpha=0.25, \beta=0.4, \gamma=0.25, K=4$ \\ $\tilde{A}=A+2I$ \\ $\overline{A}_{(i,j)}=\frac{\tilde{A}_{(i,j)}}{\sum_{(i,j)\in E}\tilde{A}_{(i,j)}+\varepsilon} \  if \   \tilde{A}_{(i,j)}\geq(\mu-\sigma)$ \\ $\overline{A}_{(i,j)}=0 \  if \   \tilde{A}_{(i,j)}\textless(\mu-\sigma)$ \\ $\mu=\frac{\sum_{(i,j)\in E}\tilde{A}_{(i,j)}}{\left|E\right|}, \sigma=\sqrt{\frac{\sum_{(i,j)\in E}\left(\tilde{A}_{(i,j)}-\mu\right)^{2}}{\left|E\right|}}$ }                         \\ \hline
\textbf{Computer}                 & \makecell[l]{$Z=\tilde{\alpha}_0X+\tilde{\alpha}_1\hat{A}XW_1+\tilde{\alpha}_2\hat{A}^2XW_2$}                            & \makecell[l]{$Initial \   \alpha=0.1, K=2$ \\ $\tilde{\alpha}_{0},\tilde{\alpha}_{1},...,\tilde{\alpha}_{K}=\frac{\alpha^{0},\alpha^{1},...,\alpha^{K}}{\sum_{k=0}^{K}\left|\alpha^{k}\right|}$  \\  $\hat{A}=\tilde{D}^{-1/2}\tilde{A}(\tilde{D}^{T})^{-1/2}, \tilde{A}=A+2I, \tilde{D}=D+2I$}                         \\ \hline
\textbf{Photo}                     & \makecell[l]{$Z_{0}=(\beta+\gamma)X$  \\  $Z_{K}=Z_{0}+\sum_{k=1}^{K}\beta\gamma\hat{A}Z_{k-1}W_{k}$}                            & \makecell[l]{ $Initial \  \gamma=0.3, \beta=0.7, K=3$ \\ $\hat{A}=\tilde{D}^{-1/2}\tilde{A}(\tilde{D}^{T})^{-1/2}, \tilde{A}=A+2I, \tilde{D}=D+2I$}                        \\ \hline
\textbf{Chameleon}                 & \makecell[l]{$Z=\tanh((XW_1+b)+\frac1{1+e^{-\alpha}}\tilde{A}X_{raw}W_2)$}                            & \makecell[l]{$Initial \   \alpha=1$  \\  $\tilde{A}=D^{-1/2}A(D^{T})^{-1/2}$}                         \\ \hline
\textbf{Squirrel}                     & \makecell[l]{ $X_{_{Att}}=softmax(X\cdot Att, dim=1)$ \\ $Z_{0}=\hat{A}(XW_{0}+b_{0})W_{1}$  \\  $Z_{1}=\hat{A}(Z_{0}W_{2}+b_{1})W_{3}$ \\ $Z=Z_{1}\cdot X_{Att}\cdot W_{3}+b_{1}$}                            & \makecell[l]{$Initial \  Att\in R^{n\times1}$ \\ $\hat{A}=\tilde{D}^{-1/2}\tilde{A}(\tilde{D}^{T})^{-1/2}, \tilde{A}=A+2I, \tilde{D}=D+2I$}                         \\ \hline
\textbf{Texas}                      & \makecell[l]{$Z=W_{2}\cdot ELU((W_{1}\cdot X_{raw}+\sum_{k=1}^{K}\tilde{A}^{k}\cdot X)\cdot Att)$}                            & \makecell[l]{$Initial \   Att\in R^{n\times1}, K=4$  \\  $\tilde{A}=D^{-1/2}A(D^{T})^{-1/2}$}                         \\ \hline
\textbf{Cornell}                     & \makecell[l]{$Z=\tilde{A}XW_{1}+\sum_{j\in N(i)}\alpha_{(i,j)}W_{2}X_{j}^{raw}$}                            & \makecell[l]{$\tilde{A}=D^{-1/2}A(D^T)^{-1/2}$}                         \\ \hline
\end{tabular}}
\end{table*}

\section{Experiments}
\subsection{Experiment Setting}
\subsubsection{Datasets}
We utilize five widely-used homophilic graphs, including the citation graphs Cora, CiteSeer, and PubMed, as well as the Amazon co-purchase graphs Computers and Photo. In addition, we employed four heterophilic benchmark datasets, including the Wikipedia graphs Chameleon and Squirrel, and the webpage graphs Texas and Cornell from WebKB. Detailed information about these datasets can be found in \textbf{Appendix A}. 
Following \cite{9,26}, for the node classification task under the transductive setting, we randomly sparse split the node set into train/val/test datasets with ratio 2.5\textbf{\%}:2.5\textbf{\%}:95\textbf{\%}. We have also included the experimental results for the randomly dense split (60\%:20\%:20\%) in \textbf{Appendix F}.

\subsubsection{Baselines}
We evaluated the performance of AutoSGNN by comparing it with several baselines: 1) \textbf{graph NAS methods}: GTFGNAS \cite{18}, F2GNN \cite{36}, Genetic-GNN \cite{37}, and SANE \cite{19}. 
2) \textbf{Spectral GNNs}: Since AutoSGNN is designed to generate spectral GNNs for specific tasks and datasets, we selected five baselines—APPNP \cite{38}, GPRGNN \cite{26}, FAGCN \cite{25}, BernNet \cite{27}, JacobiConv\cite{JacobiConv} and NFGNN \cite{9}—along with two classic spectral GNNs, GCN \cite{24} and ChebNet \cite{31}.

\subsubsection{Reproducibility}
For AutoSGNN, the iterative search algebra is set to 30, and the evolutionary strategy prompts for $E1$ and $E2$ both contain $P1=P2=4$ elite individuals. The GPT-3.5-turbo pre-trained LLM is used, with the parallel response number set to 4. Through parallel experiments with 3 prompts, the number of candidate individuals generated per cycle is 12. We use node classification accuracy on the validation dataset as the fitness metric, saving the top 30 individuals per cycle as elite individuals. We employ 2-layer generated spectral GNN layers and use 2-layer MLP for feature transformation. Similar to other spectral GNNs baselines, we follow the experimental setup in \cite{9} and use the best hyperparameter combinations provided in the original papers for each dataset. For GNN-NAS baselines, we follow the experimental setup in \cite{18}, and like AutoSGNN, we set the population size and iterative search algebra to 30. To prevent overfitting during the evaluator's training for each SGNN, we set an early stopping criterion of 200 epochs.

All experiments, in order to remove the effects of randomness, are run independently for 10 trials and the resulting means and standard deviations are reported. All search experiments are conducted on a machine equipped with four NVIDIA GeForce RTX 3090 GPUs, two Intel(R) Xeon(R) Silver 4210 CPUs (2.20 GHz), and 252GB of RAM. 

\begin{figure}[t]
\centering
\includegraphics[width=1\columnwidth]{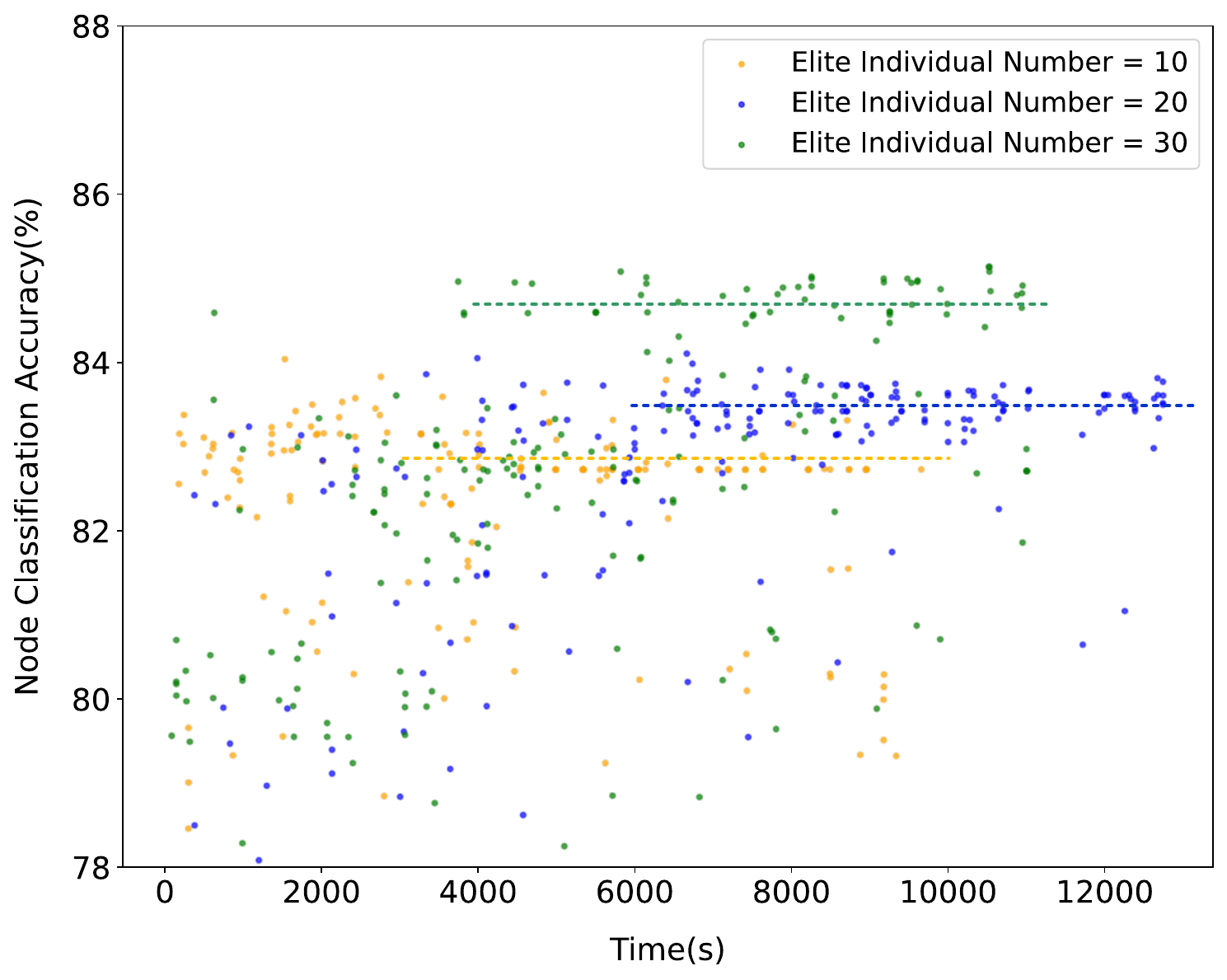} 
\caption{The iterative search process of AutoSGNN with different Elite Individual numbers for node classification experiment on the PubMed dataset. Each dot represents the node classification result of a searched condidate individual on the PubMed dataset. The dashed line represents the overall convergence trend of elite individuals.}
\label{Elite_nums}
\end{figure}

\subsection{AutoSGNN v.s. Human-Designed GNN}
To demonstrate the effectiveness of AutoSGNN, we validate the network architecture obtained through the search process using sparse node classification tasks. In Table \ref{tab:table 2}, compared to the GNN-NAS methods, AutoSGNN outperforms in most datasets. For the nine datasets, seven of them are ranked first or second best result, except for the Citeseer and Chameleon datasets. For advanced manually designed spectral GNNs, it is observed that they do not always perform well on every dataset. For instance, GPRGNN achieved the second-best node classification accuracy on the PubMed dataset but performed poorly on the Squirrel dataset. This indicates that different network architectures may have their own limitations. Selecting different network architectures for different situation can better adapt to various graph datasets.

\subsection{Case Study on Cora}
As shown in Table \ref{tab:table sum}, AutoSGNN can search for suitable spectral GNNs for different datasets and provide corresponding hyperparameters. More detailed information can be found in \textbf{Appendix C}. For the Cora dataset, the network generated by AutoSGNN can be regarded as a full-frequency learning filter composed of multiple filters. When $k=0, Z_0=\alpha X$, the initial graph features $X$ are learned, which can be considered as a low-frequency filter. When $k=K$, in the equation $Z_K=(1-\alpha)^K\hat{A}Z_{K-1}W_K + Z_{K-1}$, the term $(1-\alpha)^K\hat{A}Z_{K-1}W_K$ can be seen as a high-frequency filter. For the intermediate layer $k$, the term $\alpha(1-\alpha)^k\hat{A}Z_{k-1}W_k$ acts as a filter learning specific frequencies. Therefore, the entire propagation mechanism can be represented as a process of learning and processing graph signals across the entire spectrum. Additionally, AutoSGNN determines the hyperparameters $K=4$ and intial $\alpha=0.15$, and the graph kernel $\hat{A}$, which determine the relevant properties in the frequency domain.

\subsection{Parameter Analysis}
This section explores the impact of various components and hyperparameters on AutoSGNN.

\begin{figure}[t]
\centering
\includegraphics[width=1\columnwidth]{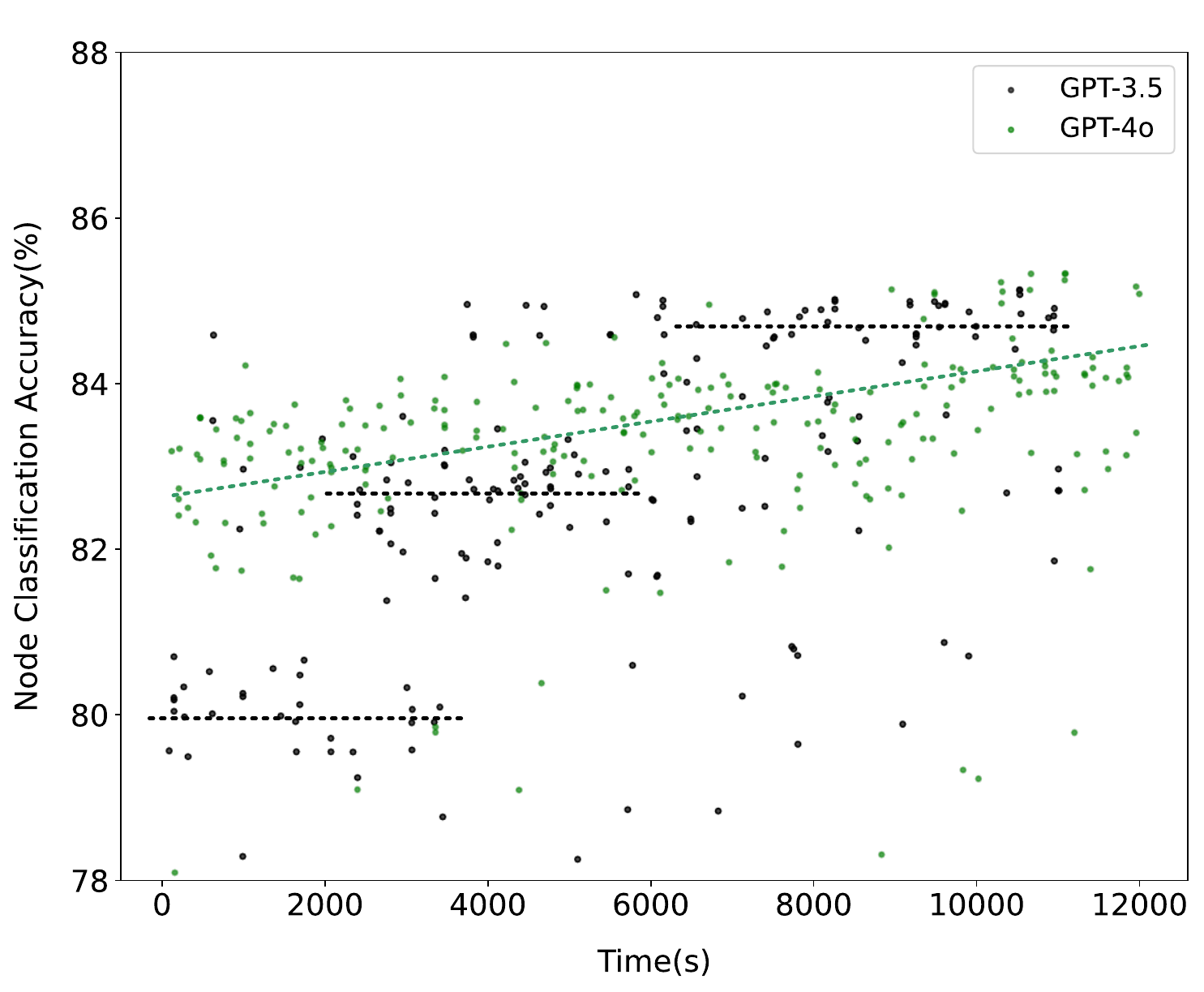} 
\caption{The iterative search process of AutoSGNN with different LLMs for the node classification experiment on the PubMed dataset.}
\label{D_LLM}
\end{figure}

\subsubsection{No. of Elite Individuals}
As shown in Figure \ref{Elite_nums}, we compared the search process for Elite Individual numbers of 10, 20, and 30. It is evident that a larger elite individual number allows for a broader search space, increasing the likelihood of finding the most suitable network architecture. Conversely, when the Elite Individual number is smaller, the population tends to converge to network architectures with generally mediocre performance.

\subsubsection{Choices of LLMs}
From Figure \ref{D_LLM}, for the PubMed dataset, the spectral GNNs generated using AutoSGNNs from ChatGPT 4o achieved a classification score of 85.43\%, about 0.3\% higher than the score obtained using ChatGPT 3.5. We can also observe that the use of more powerful LLMs in AutoSGNN results in a more stable search process. By utilizing ChatGPT-4o, the prompts' requirements are better understood, enabling a more accurate connection between the various preferences of each dataset, thus facilitating the search for an appropriate spectrum GNN. Additionally, by comparing the number of points in different colors in the figure, we can see that the use of more advanced LLMs significantly enhances the executability of the response code.

\begin{table}[t]
\centering
\caption{The accuracy of AutoSGNN and its variants on the PubMed dataset, where w.o. means remove prompt of this type.}
\label{tab:table ablation}
\begin{tabular}{c|cccc}
\hline
\textbf{AutoSGNN} & \textbf{w.o. E1} & \textbf{w.o. E2} & \textbf{w.o. C1} \\ \hline
\textbf{85.13±0.26}   & 84.88±0.43       & 84.54±0.59       & 83.95±0.67       \\ \hline
\end{tabular}
\end{table}

\subsubsection{Number of Elite Individuals in Prompt Operators.}

\begin{figure}[t]
\centering
\includegraphics[width=1.0\columnwidth]{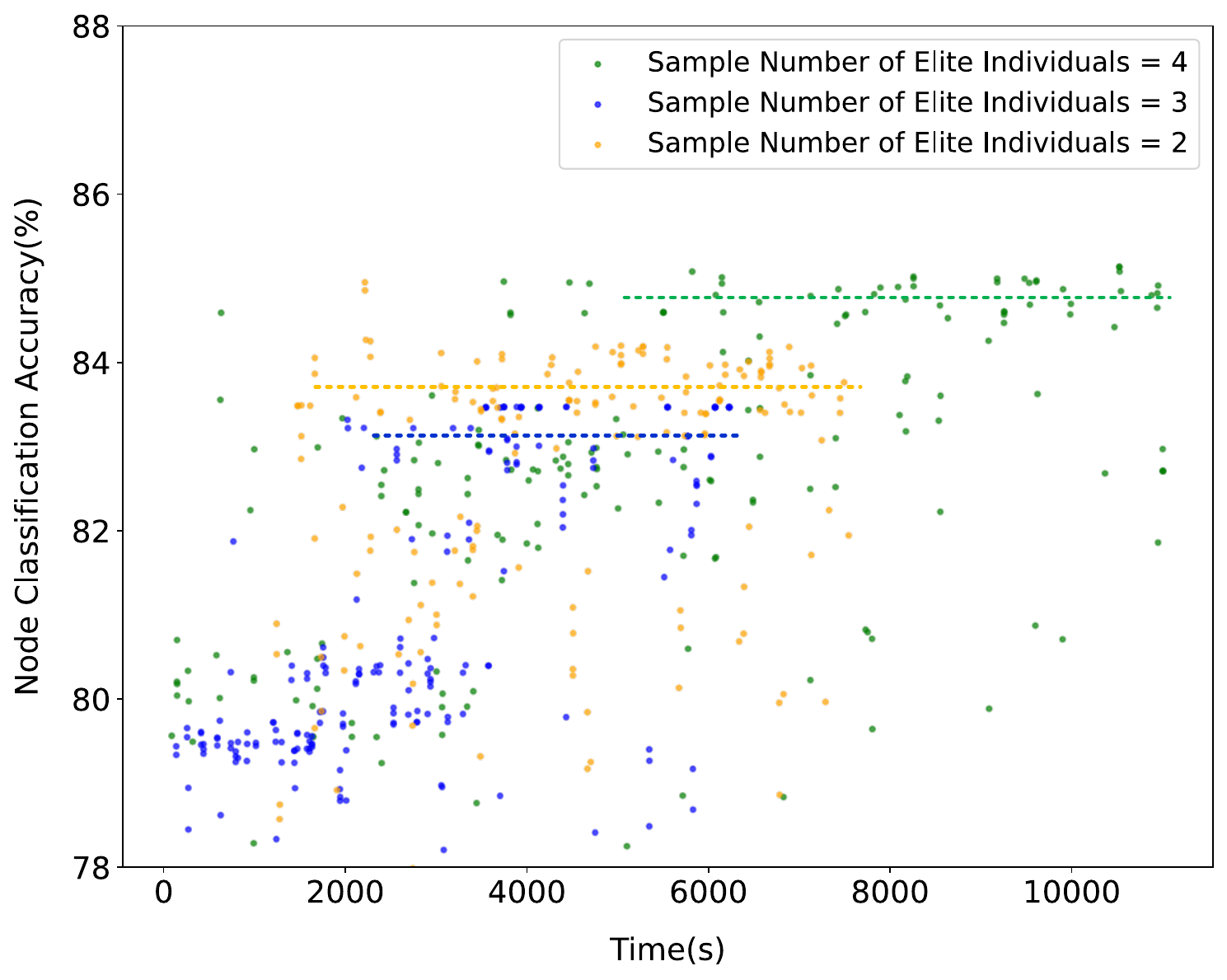} 
\caption{The iterative search process of AutoSGNN with different number of elite individuals in prompt operators for node classification experiment on the PubMed dataset. Each dot represents the node classification result of a searched condidate individual on the PubMed dataset. The dashed line represents the overall convergence trend of elite individuals.}
\label{operating_num}
\end{figure}

We explore the impact of $P1$ and $P2$, extracted per iteration by the evolutionary prompts ($E1$/$E2$) on AutoSGNN. We set the number of individuals for the operations to $P1=P2=2, 3$ and $4$, respectively. The experimental results are shown in Figure \ref{operating_num}. From the figure we can see that when $P1=P2=4$, the entire search process of AutoSGNN is more effective. When $P1=P2=2$, the performance is the second best, and when $P1=P2=3$, the performance is the worst. We believe this may be because when $P1=P2=2$, all types of prompts can be considered as preference-setting prompts ($C1$), leading to better search effectiveness compared to when $P1=P2=3$. However, relying solely on preference-setting prompts ($C1$) is not sufficient. As shown in Table \ref{tab:table ablation}, the absence of evolutionary prompts ($E1/E2$) still impacts the performance of AutoSGNN.

\subsection{Ablation Study}

We explore the impact of the three types of prompts on the AutoSGNN search process. Since removing one type of prompt causes AutoSGNN to generate only 8 candidate individuals per iteration, we set the number of iterations to 45 to ensure a fair evaluation, thereby generating the same number of candidate individuals.  The results in Table \ref{tab:table ablation} show that removing any type of prompt strategy affects the search performance of AutoSGNN. Particularly, the removal of the preference-setting prompt ($C1$) has the most significant weakening effect. This may be because without the preference-setting prompt ($C1$), the LLMs cannot quickly and effectively align with task preferences, leading to many ineffective searches.

\subsection{Time Complexity Analysis}

To compare the search efficiency of GNN-NAS methods and AutoSGNN, as shown in Figure \ref{model_running_times}, we recorded the time required for each method to complete a full run under the experimental settings described earlier.
In Figure \ref{model_running_times}, AutoSGNN is competitive. More detailed information can be found in \textbf{Appendix E}.

\subsection{Generalization of AutoSGNN}

In our exploration of spectral GNNs generated by AutoSGNN, we examined their generalization capability across different types of graph. The result in \textbf{Appendix D} illustrates that SGNNs optimized for a specific dataset generally perform the best on that dataset. However, when these SGNNs are applied to different types of graphs, they frequently fail to achieve optimal performance. This finding highlights a limitation in the generalizability of SGNNs: Their effectiveness is often contingent upon the alignment between their information processing mechanisms and the specific information requirements of downstream tasks. Thus, automatically tailoring SGNNs to match the characteristics of different graph types is advisable. For further details, please refer to \textbf{Appendix D}.

\section{Conclusion and Future Work}
This paper proposes a spectral GNNs generation method (AutoSGNN), which leverages the synergy between LLMs and ES. By utilizing three types of prompt, it sets the preferences of LLMs to adaptively search for spectral GNNs suitable for different graph. AutoSGNN is applied to nine widely used graph datasets, and the experimental results show that the spectral GNNs generated by AutoSGNN are more suitable than advanced manually designed spectral GNNs. Compared to GNN-NAS methods, AutoSGNN outperforms in both accuracy and time efficiency. Notably, unlike GNN-NAS which can be viewed as a combinatorial optimization problem, AutoSGNN is a ture automatic propagation mechanism generation method assisted by LLMs. 

In summary, we have developed a tailored search space for spectral GNNs specifically designed for LLMs. By incorporating ES, we enable LLMs to effectively learn and utilize SGNNs. This novel approach enhances the ability of LLMs to perceive, analyze, and describe graphs, offering a new perspective on the construction of graph foundation models. In future work, we plan to explore further applications of LLMs in various graph-based tasks and investigate the use of AutoSGNNs with real-world graph data.

\begin{figure}[t]
\centering
\includegraphics[width=1.0\columnwidth]{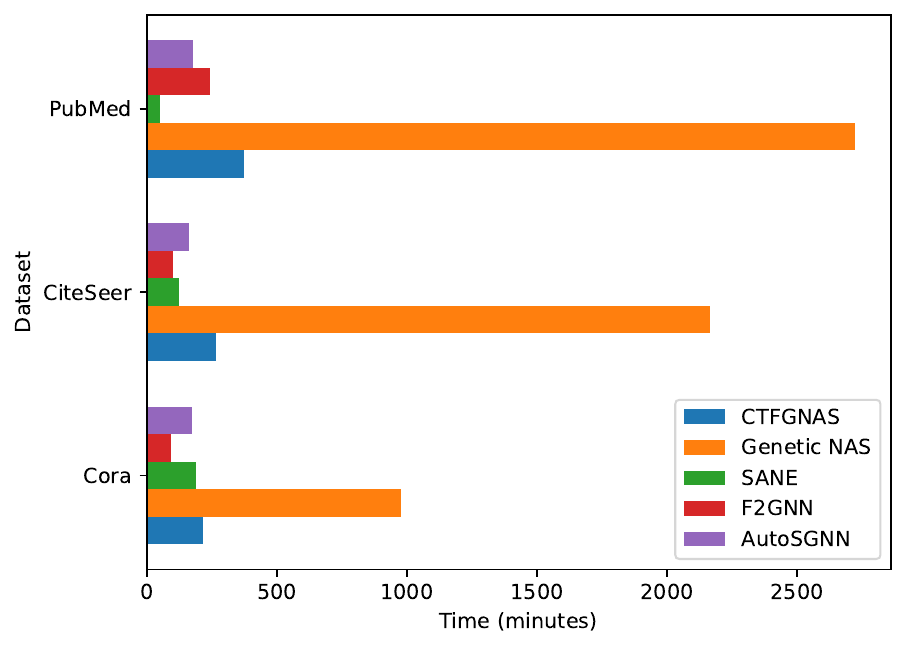} 
\caption{ Computational time (clock time in minutes) of GNN-NAS methods and AutoSGNN tested on the Pubmed dataset.}
\label{model_running_times}
\end{figure}

\section{Acknowledgments}
This work was supported in part by the National Natural Science Foundation of China under Grants 62471371, 62306224 and 62206205, in part by the Young Talent Fund of Association for Science and Technology in Shaanxi, China under Grant 20230129, in part by the Science and Technology Program of Guangzhou, China under Grant 2023A04J1060, in part by the Guangdong High-level Innovation Research Institution Project under Grant 2021B0909050008, in part by the Guangzhou Key Research and Development Program under Grant 202206030003, and in part by the Fundamental Research Funds for the Central Universities under Grant XJSJ23097.

\bibliography{aaai25}

\newpage
\setcounter{table}{0}   
\setcounter{figure}{0}
\renewcommand{\thetable}{A\arabic{table}}
\renewcommand{\thefigure}{A\arabic{figure}}
\onecolumn
\section{APPENDIX}


\subsection{A. The detailed information of datasets.}

We used five homogeneous graph datasets and four heterogeneous graph datasets from the PyTorch Geometric library\footnote{\url{https://pytorch-geometric.readthedocs.io/en/latest/cheatsheet/data_cheatsheet.html}}. Specific details of each dataset can be found in Table \ref{tab:table A1}. The five homogeneous graph datasets include citation graphs Cora, CiteSeer, PubMed, and Amazon co-purchase graphs Computers and Photo. The four heterogeneous graphs include the Wikipedia graphs Chameleon and Squirrel, as well as the social webpage connection graphs Texas and Cornell.

\begin{table}[h]
\centering
\caption{Benchmark dataset properties and statistics.}
\label{tab:table A1}
\begin{tabular}{c|ccccccccc}
\hline
\textbf{Dataset} & \textbf{Cora} & \textbf{CiteSeer} & \textbf{PubMed} & \textbf{Computers} & \textbf{Photo} & \textbf{Chameleon} & \textbf{Squireel} & \textbf{Texas} & \textbf{Cornell} \\ \hline
\textbf{Class}   & 7             & 6                 & 5               & 10                 & 8              & 5                  & 5                 & 5              & 5                \\
\textbf{Feature} & 1433          & 3703              & 500             & 767                & 745            & 2325               & 2089              & 1703           & 1703             \\
\textbf{Nodes}   & 2708          & 3327              & 19717           & 13752              & 7650           & 2277               & 5201              & 183            & 183              \\
\textbf{Edges}   & 5278          & 4552              & 44324           & 245861             & 119081         & 31371              & 198353            & 279            & 277              \\ \hline
\end{tabular}
\end{table}

\subsection{B. Evolutionary Prompts.}

\begin{figure}[!h]
\centering
\includegraphics[width=1.0\columnwidth]{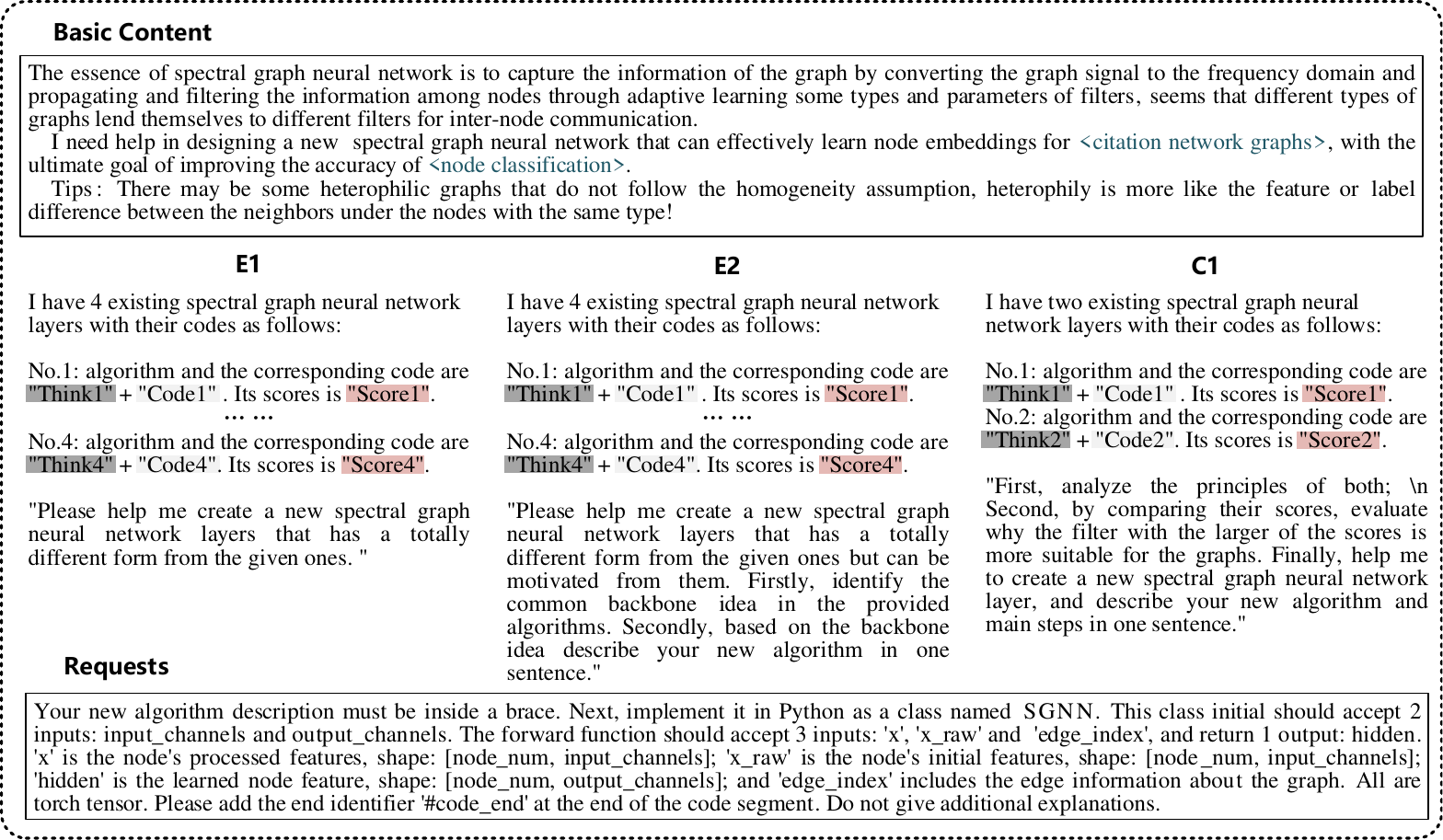} 
\caption{The detailed information of evolutionary prompts($E1$, $E2$, $C1$).}
\label{prompts_detail}
\end{figure}

As shown in Figure \ref{prompts_detail}, this paper utilizes three designed evolutionary prompts ($E1$, $E2$, $C1$) to guide the preferences of LLMs for different graph data. This approach aids in searching for and generating suitable spectral GNNs. 

The evolutionary prompts ($E1$ and $E2$) function similarly to mutation and crossover operations in evolutionary algorithms, continuously updating the elite population to gradually optimize the solution quality. $C1$ models algorithms like DPO, mimicking the contrastive learning approach with positive and negative samples. This enables LLMs to better understand the preferences of different graph data and tasks.

\subsection{C. AutoSGNN Responses to different datasets.}
For node classification on different graph datasets, AutoSGNN is used to search for and generate spectral GNNs. The detailed results obtained are reported as follows. Their corresponding mathematical expressions are shown in Table \ref{tab:table 1}.

\begin{table*}[h]
\centering
\caption{The summary results of the spectral GNNs searched by AutoSGNN for different graphs.}
\label{tab:table 1}
\resizebox{1\columnwidth}{!}{
\begin{tabular}{c|c|c}
\hline
\textbf{Graph}     & \textbf{Propagation Mechanism} & \textbf{Relevant Parameter} \\ \hline
\textbf{Cora}                 & \makecell[l]{$Z_{0}=\alpha X$   \\ $Z_{k}=Z_{k}+\sum_{k=1}^{K-1}\alpha(1-\alpha)^{k}\hat{A}Z_{k-1}W_{k}$ \\ $Z_{K}=(1-\alpha)^{K}\hat{A}Z_{K-1}W_{K}+Z_{K-1}$ }                            & \makecell[l]{$Initial\   \alpha=0.15, K=4$ \\ $\hat{A}=\tilde{D}^{-1/2}\tilde{A}(\tilde{D}^{T})^{-1/2}, \tilde{A}=A+4I, \tilde{D}=D+4I$}           \\ \hline
\textbf{Citeseer}               &  \makecell[l]{$Z_{0}=Att\cdot X$ \\ $Z_{K}=Z_{0}+\sum_{k=1}^{K-1}(Att\cdot\hat{A}Z_{k-1}W_{k}+0.2X_{raw})$}                             & \makecell[l]{$Initial Att\in R^{1\times n}, K=4$ \\ $\hat{A}=\tilde{D}^{-1/2}\tilde{A}(\tilde{D}^{T})^{-1/2}, \tilde{A}=A+2I, \tilde{D}=D+2I$ }                        \\ \hline
\textbf{PubMed}                   & \makecell[l]{$Z_{0}=\alpha X+\gamma X_{raw}$  \\  $Z_{K}=Z_{0}+\sum_{k=1}^{K}\beta\cdot ReLU(\overline{A}Z_{k-1}W_{k})$}                          & \makecell[l]{$Initial\  \alpha=0.25, \beta=0.4, \gamma=0.25, K=4$ \\ $\tilde{A}=A+2I$ \\ $\overline{A}_{(i,j)}=\frac{\tilde{A}_{(i,j)}}{\sum_{(i,j)\in E}\tilde{A}_{(i,j)}+\varepsilon} \  if \   \tilde{A}_{(i,j)}\geq(\mu-\sigma)$ \\ $\overline{A}_{(i,j)}=0 \  if \   \tilde{A}_{(i,j)}\textless(\mu-\sigma)$ \\ $\mu=\frac{\sum_{(i,j)\in E}\tilde{A}_{(i,j)}}{\left|E\right|}, \sigma=\sqrt{\frac{\sum_{(i,j)\in E}\left(\tilde{A}_{(i,j)}-\mu\right)^{2}}{\left|E\right|}}$ }                         \\ \hline
\textbf{Computer}                 & \makecell[l]{$Z=\tilde{\alpha}_0X+\tilde{\alpha}_1\hat{A}XW_1+\tilde{\alpha}_2\hat{A}^2XW_2$}                            & \makecell[l]{$Initial \   \alpha=0.1, K=2$ \\ $\tilde{\alpha}_{0},\tilde{\alpha}_{1},...,\tilde{\alpha}_{K}=\frac{\alpha^{0},\alpha^{1},...,\alpha^{K}}{\sum_{k=0}^{K}\left|\alpha^{k}\right|}$  \\  $\hat{A}=\tilde{D}^{-1/2}\tilde{A}(\tilde{D}^{T})^{-1/2}, \tilde{A}=A+2I, \tilde{D}=D+2I$}                         \\ \hline
\textbf{Photo}                     & \makecell[l]{$Z_{0}=(\beta+\gamma)X$  \\  $Z_{K}=Z_{0}+\sum_{k=1}^{K}\beta\gamma\hat{A}Z_{k-1}W_{k}$}                            & \makecell[l]{ $Initial \  \gamma=0.3, \beta=0.7, K=3$ \\ $\hat{A}=\tilde{D}^{-1/2}\tilde{A}(\tilde{D}^{T})^{-1/2}, \tilde{A}=A+2I, \tilde{D}=D+2I$}                        \\ \hline
\textbf{Chameleon}                 & \makecell[l]{$Z=\tanh((XW_1+b)+\frac1{1+e^{-\alpha}}\tilde{A}X_{raw}W_2)$}                            & \makecell[l]{$Initial \   \alpha=1$  \\  $\tilde{A}=D^{-1/2}A(D^{T})^{-1/2}$}                         \\ \hline
\textbf{Squirrel}                     & \makecell[l]{ $X_{_{Att}}=softmax(X\cdot Att, dim=1)$ \\ $Z_{0}=\hat{A}(XW_{0}+b_{0})W_{1}$  \\  $Z_{1}=\hat{A}(Z_{0}W_{2}+b_{1})W_{3}$ \\ $Z=Z_{1}\cdot X_{Att}\cdot W_{3}+b_{1}$}                            & \makecell[l]{$Initial \  Att\in R^{n\times1}$ \\ $\hat{A}=\tilde{D}^{-1/2}\tilde{A}(\tilde{D}^{T})^{-1/2}, \tilde{A}=A+2I, \tilde{D}=D+2I$}                         \\ \hline
\textbf{Texas}                      & \makecell[l]{$Z=W_{2}\cdot ELU((W_{1}\cdot X_{raw}+\sum_{k=1}^{K}\tilde{A}^{k}\cdot X)\cdot Att)$}                            & \makecell[l]{$Initial \   Att\in R^{n\times1}, K=4$  \\  $\tilde{A}=D^{-1/2}A(D^{T})^{-1/2}$}                         \\ \hline
\textbf{Cornell}                     & \makecell[l]{$Z=\tilde{A}XW_{1}+\sum_{j\in N(i)}\alpha_{(i,j)}W_{2}X_{j}^{raw}$}                            & \makecell[l]{$\tilde{A}=D^{-1/2}A(D^T)^{-1/2}$}                         \\ \hline
\end{tabular}}
\end{table*}

\newpage

\begin{figure}[!h]
\centering
\includegraphics[width=1.0\columnwidth]{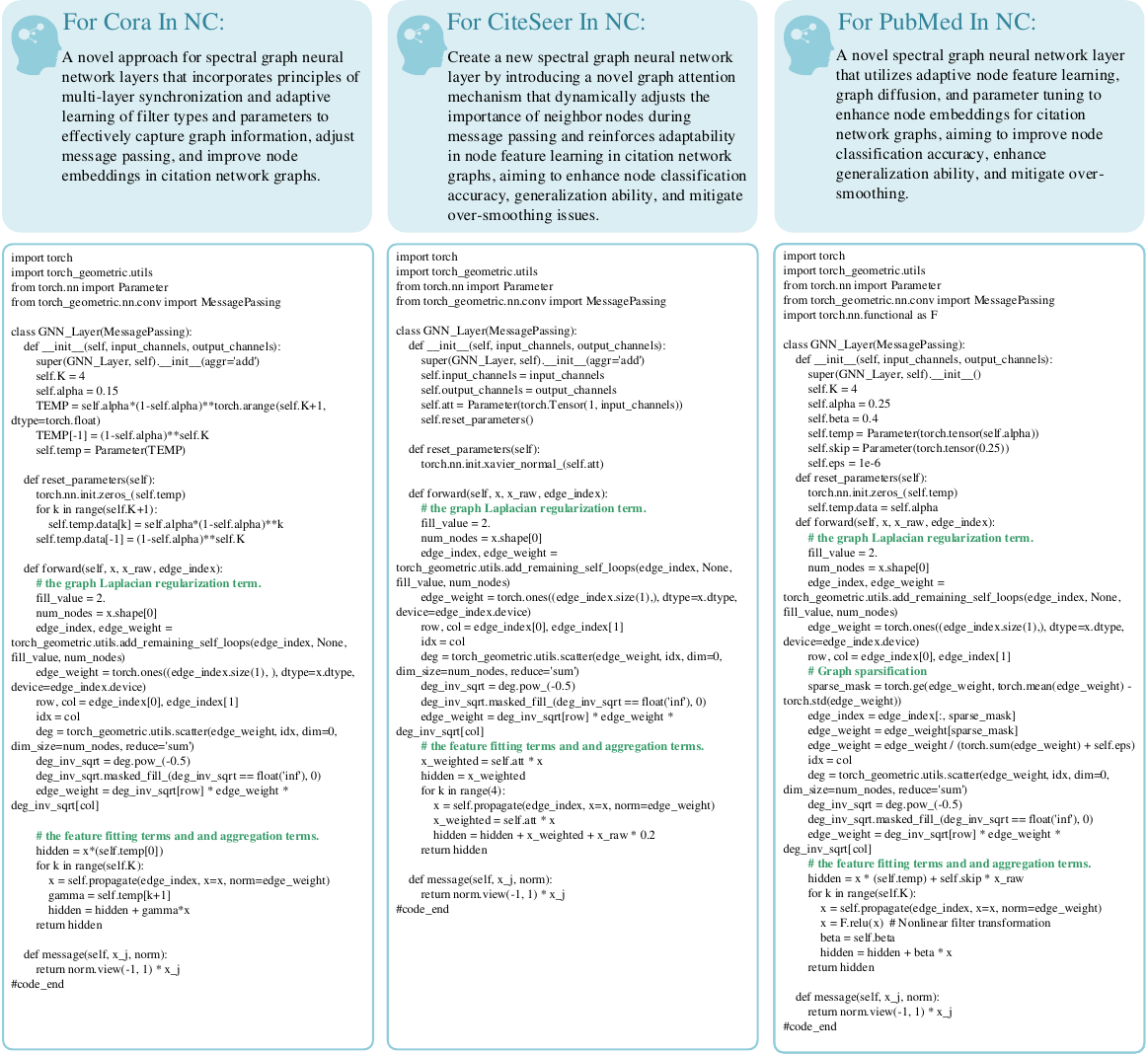} 
\caption{The optimal spectral GNNs architecture searched by AutoSGNN for node classification tasks on Cora/CiteSeer/PubMed datasets.}
\label{AutoSGNN_Respond1}
\end{figure}

\newpage

\begin{figure}[!h]
\centering
\includegraphics[width=1.0\columnwidth]{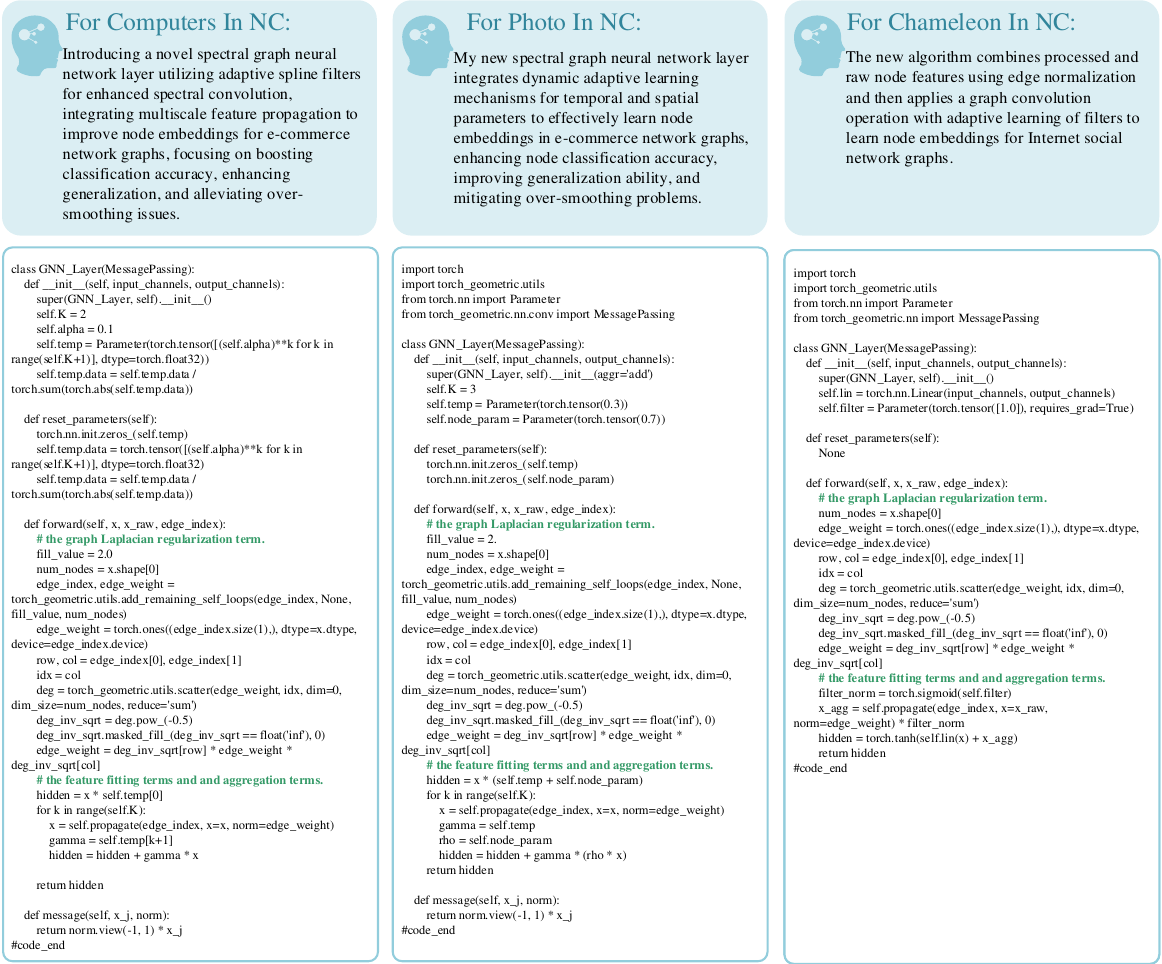} 
\caption{The optimal spectral GNNs architecture searched by AutoSGNN for node classification tasks on Computers/Photo/Chameleon datasets.}
\label{AutoSGNN_Respond2}
\end{figure}

\newpage

\begin{figure}[!h]
\centering
\includegraphics[width=1.0\columnwidth]{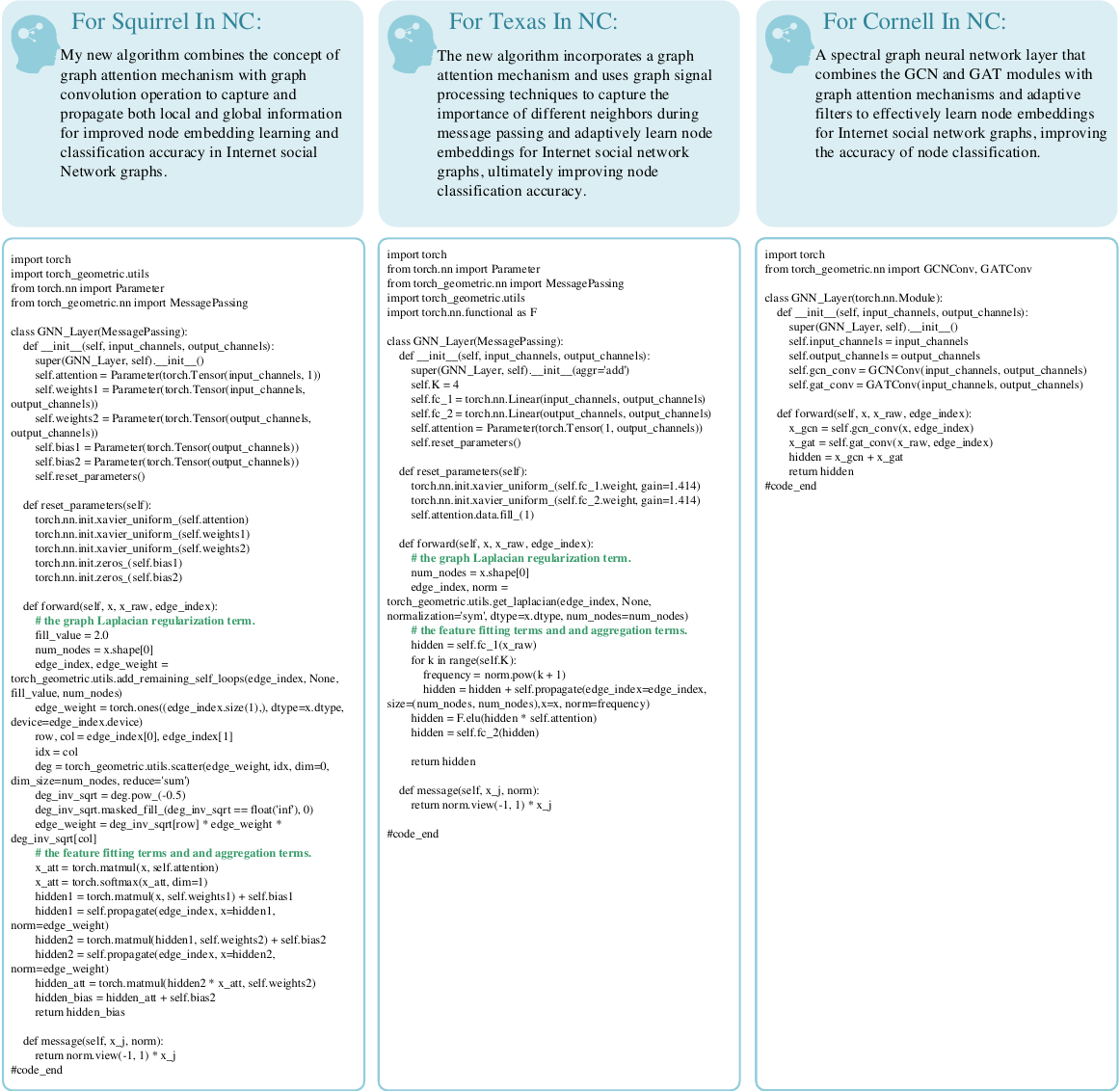} 
\caption{The optimal spectral GNNs architecture searched by AutoSGNN for node classification tasks on Squirrel/Texas/Cornell datasets.}
\label{AutoSGNN_Respond3}
\end{figure}

\newpage

\begin{figure}[!t]
\centering
\includegraphics[width=1.0\columnwidth]{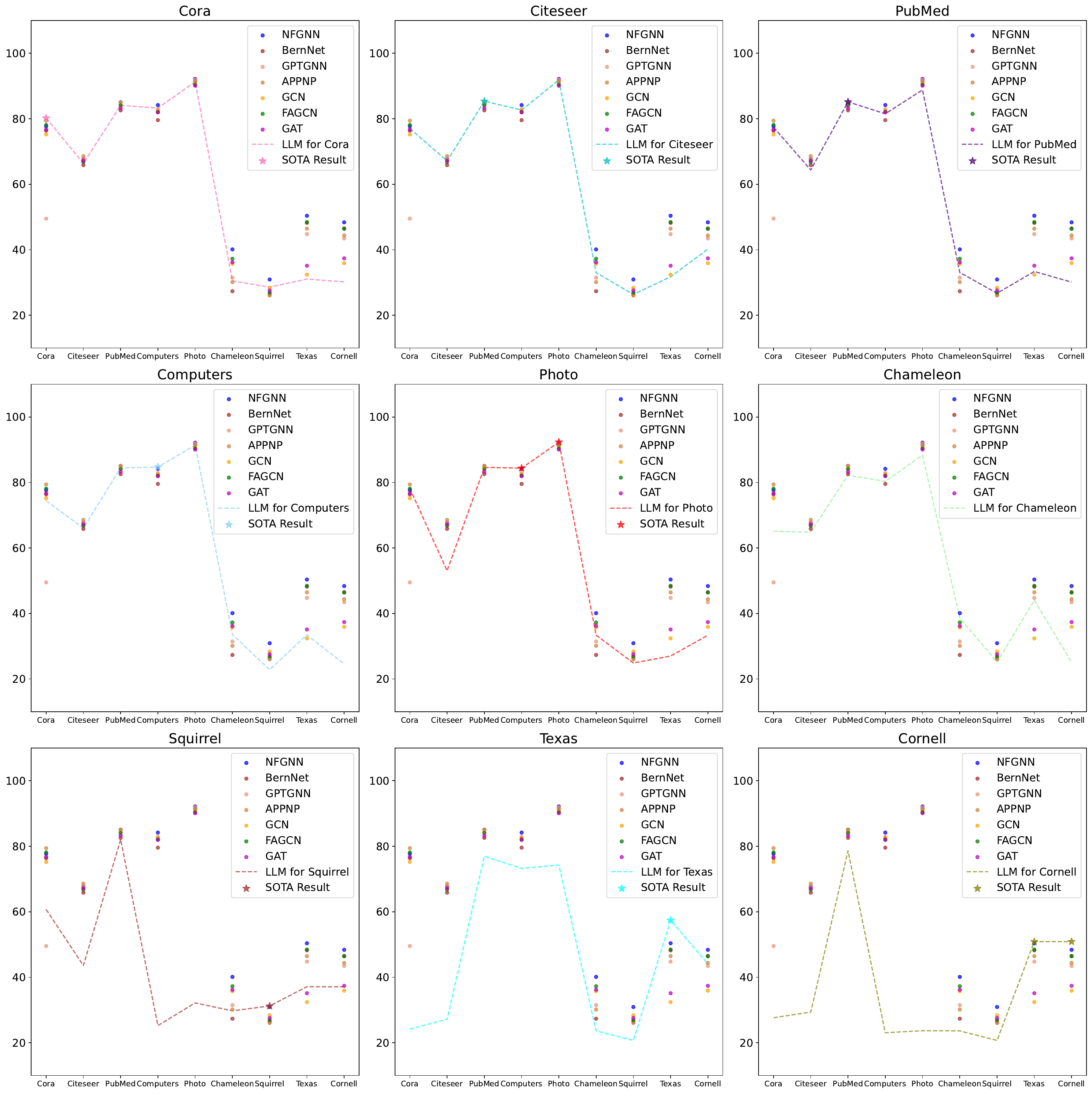} 
\caption{A spectral GNNs generated for one graph is migrated to other graphs.}
\label{RunInOther}
\end{figure}

\newpage

\subsection{D. Generalization and Migration Experiments.}


To explore the robustness of the spectral GNNs generated by AutoSGNN, we transferred the architecture generated for one graph to other graphs. The experimental results are shown in Figure \ref{RunInOther}. The title of each subplot represents the architecture generated by AutoSGNN for the node classification task on that dataset. The vertical axis indicates the node classification accuracy, while the horizontal axis represents other datasets. A star symbol denotes the optimal resolution achieved.

Each subplot represents the transfer of the spectral GNN architecture generated for the dataset mentioned in the title to other datasets. An interesting phenomenon can be observed from the figure: the spectral GNN generated for CiteSeer achieved the best classification results when transferred to PubMed; the network generated for the Photo dataset achieved the best results when transferred to Computers; and the network generated for the Cornell dataset achieved the best results when transferred to Texas. We found that these datasets belong to the same type of graph data, which may indicate that there is a certain disparity between different types of graph data. Each spectral GNN has its capability boundary, and its performance may be diminished when the test data crosses domains.

\subsection{E. Time Complexity Analysis}

To compare the search efficiency of GNNs-NAS methods and AutoSGNN, as shown in Figure \ref{model_running_times}, we recorded the time required for each method to complete a full run under the experimental settings described earlier. From the figure, we can see the following: (1) SANE and F2GNN have shorter computation times because both are differential architecture search methods. They use gradient descent search, allowing them to quickly approach a better solution. (2) In contrast, Genetic-GNN and GTFGNAS are NAS methods based on evolutionary algorithms, which consume more time overall. By comparing these methods, it can be seen that the search times for AutoSGNN on different datasets, such as Cora, Citeseer, and PubMed, are approximately 173 minutes, 160 minutes, and 176 minutes, respectively, showing little variation. The response times for the LLM are 79 minutes, 92 minutes, and 107 minutes, respectively. Due to network latency and other factors, the theoretical search time for AutoSGNN could be even lower. Therefore, overall, the search efficiency of AutoSGNN is acceptable, positioned between gradient descent NAS methods and evolutionary computation methods.

\subsection{F. Node classification tasks under the dense splitting ratios of 60\%:20\%:20\%}

Specifically, the sparse splitting ratio (2.5\%:2.5\%:95\%) is used for the semi-supervised learning setting, and the dense splitting ratio (60\%:20\%:20\%) for the full-supervised learning setting. The following are the results of node classification experiments under the  dense splitting of the graph dataset. From the results, it can be seen that AutoSGNN proves to be effective and competitive in this setup.

\begin{table*}[h]
\centering
\caption{The accuracy of node classification under the dense splitting ratios of 60\%:20\%:20\%, with \textbf{the best results} highlighted in bold black and \underline{the second-best results} underlined. The results are ranked by employing Wilcoxon-Holm analysis \cite{IsmailFawaz2018deep} at a significance level of $p=0.05$. }
\label{tab:table A3}
\resizebox{1.0\columnwidth}{!}{
\begin{tabular}{cccccc|cccc|c}
\hline
\textbf{}            & \textbf{Cora}       & \textbf{Citeseer}   & \textbf{PubMed}     & \textbf{Computer}   & \textbf{Photo}      & \textbf{Chameleon}  & \textbf{Squirrel}   & \textbf{Texas}      & \textbf{Cornell}  & \textbf{Rank}   \\ \hline
\textbf{APPNP}       & 88.20±0.62    & 80.22±0.63 & 87.98±0.67          & 88.04±0.51          & 90.32±0.17          & 51.91±0.56          & 34.77±0.34          & 91.18±0.70          & 91.80±0.63        & \textbf{6}    \\
\textbf{GPRGNN}      & 88.57±0.69       &  80.13±0.67     & 88.92±0.68  &  87.20±0.47      &  93.87±0.42     &  67.48±0.40        &  49.93±0.53         &   92.92±0.61        &  91.36±0.70       & \textbf{4}   \\
\textbf{FAGCN}       &  88.57±0.92   & \textbf{83.51±0.43}      &  86.08±0.33         &  89.68±0.97         & 93.67±0.50          & 61.59±1.98          & 44.41±0.62          &  89.61±1.52         &  88.52±1.33       & \textbf{5}  \\
\textbf{BernNet}     & 88.06±0.91          & 80.17±0.78         &  88.79±0.25        &  88.61±0.41         & 93.32±0.40         & \underline{68.73±0.57}      &  50.75±0.67       &  92.30±1.23       & \underline{91.96±1.07}      & \textbf{3}   \\
\textbf{NFGNN}       &  \textbf{89.82±0.43}    & 80.56±0.55    & \underline{89.89±0.68}   & \textbf{ 90.31±0.42 }          & \underline{94.85±0.24}    & \textbf{72.52±0.59}    &  \textbf{58.90±0.35}        &  \textbf{94.03±0.82}       &  91.90±0.91        & \textbf{1}   \\
\textbf{GCN}         &  86.99±1.23        & 79.67±0.86      &  86.65±0.70        &  86.64±1.04        & 92.49±0.55         &  60.96±0.78       &  45.66±0.39       &  75.16±0.96       & 66.72±1.37    & \textbf{8}   \\
\textbf{ChebNet}     & 86.36±0.46       &  79.32±0.39        &  88.12±0.43       & 87.82±0.72         &  93.58±0.19        & 59.96±0.51         &  40.67±0.31       & 86.08±0.96        & 85.33±1.04       & \textbf{7}   \\ \hline
\textbf{AutoSGNN}    & \underline{88.77±0.80}   & \underline{81.50 ±1.53}      & \textbf{91.97±0.35} & \underline{90.06±0.63 }   &  \textbf{95.12±0.37} &  66.30±1.94       &  \underline{ 56.58±1.14}   &  \underline{93.77±1.77} & \textbf{92.95±3.95}  & \textbf{2} \\ \hline
\end{tabular}}
\end{table*}

\end{document}